\pdfoutput=1
\documentclass[11pt,a4paper]{article}
\usepackage[hyperref]{naaclhlt2019}
\usepackage{times}
\usepackage{latexsym}
\usepackage{todonotes}
\usepackage{diagbox}
\usepackage{url}

\aclfinalcopy 

\title{Harvey Mudd College at SemEval-2019 Task 4: The 
Clint Buchanan Hyperpartisan News Detector}

\author{Mehdi Drissi, Pedro Sandoval, Vivaswat Ojha, and Julie Medero \\
  Harvey Mudd College, CA \\
  \{\tt mdrissi,  psandovalsegura, vmojha, jmedero\}@hmc.edu
}

\date{}

\begin{document}
\maketitle

\begin{abstract}
  We investigate the recently developed Bidirectional Encoder Representations from Transformers (BERT) model \cite{Devlin2018} for the hyperpartisan news detection task. Using a subset of hand-labeled articles from SemEval as a validation set, we test the performance of different parameters for BERT models. We find that accuracy from two different BERT models using different proportions of the articles is consistently high, with our best-performing model on the validation set achieving 85\% accuracy and the best-performing model on the test set achieving 77\%. We further determined that our model exhibits strong consistency, labeling independent slices of the same article identically. Finally, we find that randomizing the order of word pieces dramatically reduces validation accuracy (to approximately 60\%), but that shuffling groups of four or more word pieces maintains an accuracy of about 80\%, indicating the model mainly gains value from local context.
\end{abstract}

\section{Introduction}
SemEval Task 4 \cite{kiesel:2019} tasked participating teams with identifying news articles that are misleading to their readers, a phenomenon often associated with ``fake news'' distributed by partisan sources \cite{DBLP:journals/corr/PotthastKRBS17}. 

We approach the problem through transfer learning to fine-tune a model for the document classification task. We use the BERT model based on the implementation of the github repository \textit{pytorch-pretrained-bert}\footnote{https://github.com/huggingface/pytorch-pretrained-BERT} on some of the data provided by Task 4 of SemEval. BERT has been used to learn useful representations for a variety of natural language tasks, achieving state of the art performance in these tasks after being fine-tuned \cite{Devlin2018}.  It is a language representation model that is designed to pre-train deep bidirectional representations by jointly conditioning on both left and right context in all layers. Thus, it may be able to adequately account for complex characteristics as such blind, prejudiced reasoning and extreme bias that are important to reliably identifying hyperpartisanship in articles.

We show that BERT performs well on hyperpartisan sentiment classification. We use unsupervised learning on the set of 600,000 source-labeled articles provided as part of the task, then train using supervised learning for the 645 hand-labeled articles. We believe that learning on source-labeled articles would bias our model to learn the partisanship of a source, instead of the article. Additionally, the accuracy of the model on validation data labeled by article differs heavily when the articles are labeled by publisher. Thus, we decided to use a small subset of the hand-labeled articles as our validation set for all of our experiments. As the articles are too large for the model to be trained on the full text each time, we consider the number of word-pieces that the model uses from each article a hyperparameter.

A second major issue we explore is what information the model is using to make decisions. This is particularly important for BERT because neural models are often viewed like black boxes. This view is problematic for a task like hyperpartisan news detection where users may reasonably want explanations as to why an article was flagged. We specifically explore how much of the article is needed by the model, how consistent the model behaves on an article, and whether the model focuses on individual words and phrases or if it uses more global understanding. We find that the model only needs a short amount of context (100 word pieces), is very consistent throughout an article, and most of the model's accuracy arises from locally examining the article.

In this paper, we demonstrate the effectiveness of BERT models for the hyperpartisan news classification task, with validation accuracy as high as 85\% and test accuracy as high as 77\% \footnote{All of our code can be found here, https://github.com/hmc-cs159-fall2018/final-project-team-mvp-10000}. We also make significant investigations into the importance of different factors relating to the articles and training in BERT's success. The remainder of this paper is organized as follows. Section~\ref{sec:relatedwork} describes previous work on the BERT model and semi-supervised learning. Section~\ref{sec:methodology} outlines our model, data, and experiments. Our results are presented in Section~\ref{sec:results}, with their ramifications discussed in Section~\ref{sec:discussion}. We close with an introduction to our system's namesake, fictional journalist Clint Buchanan, in Section~\ref{sec:namesake}. 

\section{Related Work}
\label{sec:relatedwork}
        
We build upon the Bidirectional Encoder Representations from Transformers (BERT) model. BERT is a deep bidirectional transformer that has been successfully tuned to a variety of tasks \cite{Devlin2018}. BERT functions as a language model over character sequences, with tokenization as described by \citet{Sennrich2015}. The transformer architecture \cite{Vaswani2017} is based upon relying on self-attention layers to encode a sequence. To allow the language model to be trained in a bidirectional manner instead of predicting tokens autoregressively, BERT was pre-trained to fill in the blanks for a piece of text, also known as the Cloze task \cite{taylor1953cloze}. 
    
Due to the small size of our training data, it was necessary to explore techniques from semi-supervised learning. \citet{Dai2015} found pre-training a model as a language model on a larger corpus to be beneficial for a variety of experiments. We also investigated the use of self-training \cite{zhu2005semi} to increase our effective training dataset size. Lastly, the motivation of examining the effective context of our classification model was based on \citet{brendel2018approximating}. It was found that  much higher performance than expected was achieved on the ImageNet dataset \cite{LiFei-Fei2009} by aggregating predictions from local patches. This revealed that typical ImageNet models could acquire most of their performance from local decisions. 

\section{Methodology}
\label{sec:methodology}

Next, we describe the variations of the BERT model used in our experiments, the data we used, and details of the setup of each of our experiments. 

\subsection{Model}

We adjust the standard BERT model for the hyperpartisan news task, evaluating its performance both on a validation set we construct and on the test set provided by Task 4 at SemEval. The training of the model follows the methodology of the original BERT paper.

We choose to experiment with the use of the two different pre-trained versions of the BERT model, \textit{BERT-LARGE} and \textit{BERT-BASE}. The two differ in the number of layers and hidden sizes in the underlying model. \textit{BERT-BASE} consists of $12$ layers and $110$ million parameters, while \textit{BERT-LARGE} consists of $24$ layers and $340$ million parameters.

\subsection{Training and Test Sets}

We focus primarily on the smaller data set of 645 hand-labeled articles provided to task participants, both for training and for validation. We take the first 80\% of this data set for our training set and the last 20\% for the validation set. Since the test set is also hand-labeled we found that the 645 articles are much more representative of the final test set than the articles labeled by publisher. The model's performance on articles labeled by publisher was not much above chance level.

Due to an intrinsic limitation of the BERT model, we are unable to consider sequences of longer than 512 word pieces for classification problems. These word pieces refer to the byte-pair encoding that BERT relies on for tokenization. These can be actual words, but less common words may be split into subword pieces \cite{Sennrich2015}. The longest article in the training set contains around $6500$ word pieces. To accommodate this model limitation, we work with truncated versions of the articles.

We use the additional $600,000$ training articles labeled by publisher as an unsupervised data set to further train the BERT model.

\subsection{Experiments}

We first investigate the impact of pre-training on \textit{BERT-BASE}'s performance. We then compare the performance of \textit{BERT-BASE} with \textit{BERT-LARGE}. For both, we vary the number of word-pieces from each article that are used in training. We perform tests with 100, 250 and 500 word pieces.

We also explore whether and how the BERT models we use classify different parts of each individual article. Since the model can only consider a limited number of word pieces and not a full article, we test how the model judges different sections of the same article. Here, we are interested in the extent to which the same class will be assigned to each segment of an article. 
Finally, we test whether the model's behavior varies if we randomly shuffle word-pieces from the articles during training. Our goal in this experiment is to understand whether the model focuses on individual words and phrases or if it achieves more global understanding. We alter the the size of the chunks to be shuffled ($N$) in each iteration of this experiment, from shuffling individual word-pieces ($N=1$) to shuffling larger multiword chunks. 

\section{Results}
\label{sec:results}

Our results are primarily based on a validation set we constructed using the last 20\% of the hand-labeled articles. It is important to note that our validation set was fairly unbalanced. About 72\% of articles were not hyperpartisan and this mainly arose because we were not provided with a balanced set of hand-labeled articles. The small validation split ended up increasing the imbalance in exchange for training on a more balanced set. The test accuracies we report were done on SemEval Task 4's balanced test dataset.

\subsection{Importance of Pre-training}
Our first experiment was checking the importance of pre-training. We pre-trained BERT-base on the 600,000 articles without labels by using the same Cloze task \cite{taylor1953cloze} that BERT had originally used for pre-training. We then trained the model on sequence lengths of 100, 250 and 500. The accuracy for each sequence length after 100 epochs is shown in \ref{table:1} and is labeled as UP (unsupervised pre-training). The other column shows how well \textit{BERT-base} trained without pre-training. We found improvements for lower sequence lengths, but not at 500 word pieces. Since the longer chunk should have been more informative, and since our hand-labeled training set only contained 516 articles, this likely indicates that BERT experiences training difficulty when dealing with long sequences on such a small dataset. As the cost to do pre-training was only a one time cost all of our remaining experiments use a pre-trained model.

{
    \begin{table}[ht]
    \centering
    \begin{tabular}{|l|l|l|}
    \hline
    Max Seq Len & BERT-base & BERT-base + UP \\ \hline
    100 & 76.7 & 79.8 \\ \hline
    250 & 75.9 & 82.9 \\ \hline
    500 & 79.1 & 75.2 \\ \hline
    \end{tabular}
    \caption{Validation accuracy for BERT-base with and without Unsupervised Pre-training (UP).}
    \label{table:1}
    \end{table}
    }

We evaluated this model on the \textit{SemEval 2019 Task 4: Hyperpartisan News Detection} competition's \textbf{pan19-hyperpartisan-news-detection-by-article-test-dataset-2018-12-07} dataset using TIRA \cite{TIRA}. Our model, with a maximium sequence length of 250, had an accuracy of $77\%$. It had higher precision ($83.2\%$) than recall ($67.8\%$), for an overall F1-score of $0.747$. 

\subsection{Importance of Sequence Length}
Next, we further explore the impact of sequence length using \textit{BERT-LARGE}. The model took approximately 3 days to pre-train when using 4 NVIDIA GeForce GTX 1080 Ti. On the same computer, fine tuning the model on the small training set took only about 35 minutes for sequence length 100. The model's training time scaled roughly linearly with sequence length. We did a grid search on sequence length and learning rate.
	
\begin{table*}[ht]
\centering
\begin{tabular}{|l|l|l|l|l|l|l|}
\hline
\diagbox{Max Seq Len}{Learning Rate} & 5e-7 & 1e-6 & 1.5e-6 & 2e-6 & 2.5e-6 & 3e-6 \\ \hline
50                           & 78.3 & 80.6 & 79.8   & 79.1 & 79.1   & 77.5 \\ \hline
100                          & 83.7 & 83.7 & 86.1   & 86.1 & 85.3   & 84.5 \\ \hline
150                          & 77.5 & 79.8 & 81.4   & 80.6 & 79.8   & 79.8 \\ \hline
200                          & 81.4 & 80.6 & 79.8   & 84.5 & 83     & 81.4 \\ \hline
\end{tabular}
\caption{Validation Accuracy on \textit{BERT-LARGE} across sequence length and learning rate.}
\label{table:2}
\end{table*}

Table~\ref{table:2} shows that the model consistently performed best at a sequence length of 100. This is a discrepancy from \textit{BERT-BASE} indicating that the larger model struggled more with training on a small amount of long sequences. For our best trained \textit{BERT-LARGE}, we submitted the model for evaluation on TIRA. Surprisingly, the test performance (75.1\%) of the larger model was worse than the base model. The experiments in \cite{Devlin2018} consistently found improvements when using the large model. The main distinction here is a smaller training dataset than in their tasks. The experiments in the remaining sections use the same hyperparameters as the optimal \textit{BERT-LARGE}.

\subsection{Model Consistency}
Due to the small training dataset, we tried self-training to increase our effective training set. We trained the model for 40 epochs. For the remaining 60 epochs, after each epoch we had the model make predictions on five slices of 500 unlabeled articles. If an article had the same prediction for more than four slices, we added it to the labeled training data. The model always added every article to the training set, though, since it always made the same prediction for all 5 slices. This caused self-training to be ineffective, but also revealed that the model's predictions were very consistent across segments of a single article.\footnote{We also tried training a model that averaged its predictions across multiple slices. This turned out to be slightly worse, likely due to the model's high consistency.}

\subsection{Effective Model Context}
Finally, we investigate whether the model's accuracy primarily arose from examining words or short phrases, or if the decisions were more global. We permuted the word pieces in the article at various levels of granularity. At the finest level (permute\_ngrams = 1), we permuted every single word piece, forcing the model to process a bag of word pieces. At coarser levels, ngrams were permuted. As the sequence length for these experiments was 100, permute\_ngrams = 100 corresponds to no permutation. The results can be found in \ref{table:3}.

\begin{table}[ht]
\centering
\begin{tabular}{|l|l|}
\hline
permute\_ngrams & Validation Accuracy \\ \hline
1   & 67.4 \\ \hline
2   & 62.8 \\ \hline
3   & 75.2 \\ \hline
4   & 83.0 \\ \hline
5   & 76.0 \\ \hline
10  & 82.2 \\ \hline
20  & 76.7 \\ \hline
50  & 79.8 \\ \hline
100 & 84.5 \\ \hline
\end{tabular}
\caption{\textit{BERT-LARGE} across permute\_ngrams.}
\label{table:3}
\end{table}

Accuracy drops a lot with only a bag of word pieces, but still reaches 67.4\%. Also, most of the accuracy of the model (within 2\%) is achieved with only 4-grams of word pieces, so the model is not getting much of a boost from global content.

\section{Discussion}
\label{sec:discussion}

Our successful results demonstrate the adaptability of the BERT model to different tasks. With a relatively small training set of articles, we were able to train models with  high accuracy on both the validation set and the test set.

Our models classified different parts of a given article identically, demonstrating that the overall hyperpartisan aspects were similar across an article. In addition, the model had significantly lower accuracy when word pieces were shuffled around, but that accuracy was almost entirely restored when shuffling around chunks of four or more word pieces, suggesting that most of the important features can already be extracted at this level.

In future work, we we would like to make use of the entire article. Naively, running this over each chunk would be computationally infeasible, so it may be worth doing a full pass on a few chunks and cheaper computations on other chunks. 

\section{Namesake}
\label{sec:namesake}

\begin{figure}[ht]
    \centering
    \includegraphics[width=.25\textwidth]{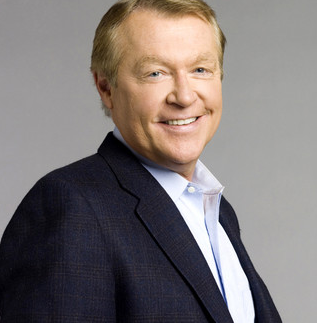}
    \caption{Jerry verDorn as Clint Buchanan.} 
    \label{fig:buchanan}
\end{figure}

Our system is named after Clint Buchanan\footnote{http://abc.go.com/shows/one-life-to-live/bio/clint-buchanan/165745}, a fictional journalist on the soap opera \textit{One Life to Live}. Following the unbelievable stories of Clint and his associates may be one of the few tasks \textit{more} difficult than identifying hyperpartisan news.

\clearpage

\bibliography{naaclhlt2019}
\bibliographystyle{acl_natbib}

\appendix

\end{document}